\documentclass{article} 
\usepackage{iclr2025_conference,times}

\usepackage{amsmath,amsfonts,bm}

\def\eqref#1{equation~\ref{#1}}

\def\1{\bm{1}}

\DeclareMathAlphabet{\mathsfit}{\encodingdefault}{\sfdefault}{m}{sl}
\SetMathAlphabet{\mathsfit}{bold}{\encodingdefault}{\sfdefault}{bx}{n}

\DeclareMathOperator*{\argmin}{arg\,min}

\usepackage{hyperref}
\usepackage{url}
\usepackage{graphicx}
\usepackage{amsmath}
\usepackage{amssymb}
\usepackage{amsthm,amsmath,amssymb}
\usepackage{subfigure}
\usepackage{algorithm}
\usepackage{algorithmic}
\usepackage{subcaption}
\usepackage{adjustbox}
\usepackage{multirow}
\usepackage{multicol}
\usepackage{mathrsfs}
\usepackage{booktabs}
\usepackage{wrapfig}

\title{Data Extrapolation for Text-to-image Generation on Small Datasets}

\author{Senmao Ye\\
	South China University of Technology\\
	\texttt{senmaoy@gmail.com}\\
	\And 
	Fei Liu\\
	South China University of Technology\\
	\texttt{feiliu@scut.edu.cn}\\
}
\begin{document}

\maketitle
\begin{abstract}
Text-to-image generation requires large amount of training data to synthesizing high-quality images. For augmenting training data, previous methods rely on data interpolations like cropping, flipping, and mixing up, which fail to introduce new information and yield only marginal improvements. In this paper, we propose a new data augmentation method for text-to-image generation using linear extrapolation.   
Specifically,  we apply linear extrapolation only on text feature, and new image data are retrieved from the internet by search engines. For the reliability of new text-image pairs, we design two outlier detectors to purify retrieved images. Based on extrapolation, we construct training samples dozens of times larger than the original dataset, resulting in a significant improvement in text-to-image performance. Moreover, we propose a NULL-guidance to refine score estimation, and apply recurrent affine transformation to fuse text information.  Our model achieves FID scores of 7.91, 9.52 and 5.00 on the CUB, Oxford and COCO datasets. The code and data will be available on GitHub.

\end{abstract}
\begin{wrapfigure}{r}{0.42\textwidth} 
  \begin{center}
    \includegraphics[width=0.42\textwidth]{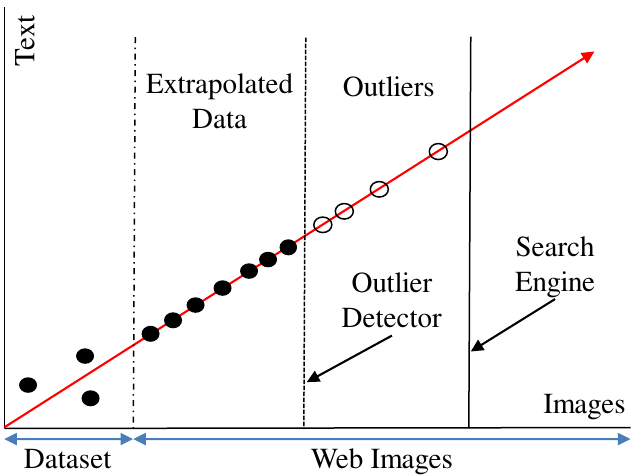}
    \caption{An illustration of data linear extrapolation. We use search engine and outlier detectors to ensure the image similarity. Extrapolation produces much more text-image pairs than the original dataset.}
    \label{fig: motivation}
  \end{center}
   \vspace{-23pt}

\end{wrapfigure}

 \section{Introduction}
Text-to-image generation aims to synthesize images according to textual descriptions. As the bridge between human language and generative models, text-to-image generation ~\citep{reed2016generative,ye2023recurrent,sauer2023stylegan,rombach2022high,ramesh2022hierarchical}is applied to more and more application domains, such as digital human~\citep{yin2023systematic}, image editing~\citep{brack2024ledits++}, and computer-aided design~\citep{liu20233dall}.
The diversity of applications leads to a large number of small datasets, where existing data are not sufficient to train high-quality generative models, and generative large models cannot overcome the long-tail effect of diverse applications.

To augment training data, existing methods typically rely on data interpolation techniques such as cropping, flipping, and mixing up images~\citep{zhang2017mixup}. While these methods leverage human knowledge to create new perspectives on existing images or features, they do not introduce new information and yield only marginal improvements. Additionally, Retrieval-base models~\citep{chen2022re,sheynin2022knn,li2022memory} employs retrieval methods to gather relevant training data from external databases like WikiImages. However, these external databases often contain very few images for specific entries, and their description styles differ significantly from those in text-to-image datasets. Furthermore, VQ-diffusion~\citep{gu2022vector} pre-trains its text-to-image model on the Conceptual Caption dataset with 15 million images, but the resulting improvements are not obvious.

In this paper, we explore data linear extrapolation to augment training data. Linear extrapolation can be risky, as similar text-image pairs may not be nearby in Euclidean space. For information reliability, as depicted in Figure~\ref{fig: motivation}, we explore linear extrapolation only on text data, and new image data are retrieved from the internet by search engines. And then outlier detectors are designed to purify retrieved web images. In this way, the reliability of new text-image pairs are guaranteed by search precision and outlier detection.

To detect outliers from web images, we divide outliers into irrelevant and similar ones. For detecting irrelevant outliers, K-means~\citep{lloyd1982least} algorithm is used to cluster noisy web images into similar images and outliers. In the image feature space generated by a CLIP encoder~\citep{radford2021learning}, similar images will be close to dataset images, while outliers will be far away. Based on this observation, we remove images that differ significantly from dataset images. For detecting similar outliers, each web image is assigned a label by a fine-grained classifier trained on the original dataset. If the label does not match the search keyword, the image is considered as an outlier and removed. For every purified web image, we extrapolate a new text descriptions according to the local manifold of dataset images. Based on extrapolation, we construct training samples dozens of times larger than the original dataset.

Moreover, we propose NULL-condition guidance to refine score estimation for text-to-image generation. Classifier-free guidance~\citep{ho2022classifier} uses a dummy label to refine label-conditioned image synthesis. Similarly, in text-to-image generation, such a dummy label can be replaced by a prompt with no new physical meaning. For example, ``a picture of bird" provides no information for the CUB dataset ``a picture of flower" provides no information for the Oxford dataset). In addition, we apply recurrent affine transformation (RAT) in the diffusion model for handling complex textual information.

The contributions of this paper are summarized as follows:

	\begin{itemize}

		\item We propose a new data augmentation method for text-to-image generation using linear extrapolation. Specifically,  we apply linear extrapolation only on text feature, and new image data are retrieved from the internet by search engines. 

		\item We propose a NULL-condition guidance to refine the score estimation for text-to-image generation. This guidance is also applicable to existing text-to-image models without further training. 
        \item  We apply recurrent affine transformation in the diffusion model for handling complex textual information.

	\end{itemize}
	\section{Related work}

\paragraph{GAN-based text-to-image models.}
Text-to-image synthesis is a key task within conditional image synthesis~\citep{DBLP:journals/tmm/FengNLW22,DBLP:journals/tmm/TanLYL22,peng2021knowledge,hou2022textface}. The pioneering work of~\citep{reed2016generative} first tackled this task using conditional GANs~\citep{DBLP:journals/corr/MirzaO14}. To better integrate text information into the synthesis process, DF-GAN~\citep{DBLP:journals/corr/abs-2008-05865} introduced a deep fusion method featuring multiple affine layers within a single block. Unlike previous approaches, DF-GAN eliminated the normalization operation without sacrificing performance, thus reducing computational demands and alleviating limitations associated with large batch sizes. Building on DF-GAN, RAT-GAN employed a recurrent neural network to progressively incorporate text information into the synthesized images. GALIP~\citep{tao2023galip} and StyleGAN-T~\citep{sauer2023stylegan} explore the potential of combining GAN models with transformers for large-scale text-to-image synthesis. However, the aforementioned GAN-based models often struggle to produce high-quality images.

\paragraph{Diffusion-based text-to-image models.}
Recently, diffusion models~\citep{ho2020denoising,DBLP:conf/nips/SongE19,DBLP:conf/iclr/0011SKKEP21,hyvarinen2005estimation} have demonstrated impressive generation performance across various tasks. Building on this success, Imagen~\citep{saharia2022photorealistic} and DALL·E 2~\citep{ramesh2022hierarchical} can synthesize images that are sufficiently realistic for real-world applications. To alleviate computational burdens, they first generate 64×64 images and then upsample them to high-resolution using another diffusion model. Additionally, the Latent Diffusion Model~\citep{rombach2022high} encodes high-resolution images into low-resolution latent codes, avoiding the exponential computation costs associated with increased resolution. DiT~\citep{peebles2023scalable} integrated latent diffusion models and transformers to enhance performance on large datasets. VQ-Diffusion~\citep{gu2022vector}  pre-train their text-to-image model on the Conceptual Caption dataset, which contains 15 million text-image pairs, and then fine-tune it on smaller datasets like CUB, Oxford, and COCO. Hence, VQ-Diffusion is the work most similar to ours but we use significantly less pre-training data while achieving better results.
\paragraph{Data augmentation methods.}
Data augmentation  increases training data to improve the performance of deep learning applications, from image classification~\citep{krizhevsky2012imagenet} to speech recognition~\citep{graves2013speech,amodei2016deep}. Common techniques include rotation, translation, cropping, resizing, flipping~\citep{lecun2015deep,vedaldi2016vgg}, and random erasing~\citep{zhong2020random} to promote visually plausible invariances. Similarly, label smoothing is widely used to boost the robustness and accuracy of trained models~\citep{muller2019does,lukasik2020does}. Mixup~\citep{zhang2017mixup} involves training a neural network on convex combinations of examples and their labels. However, interpolated samples fail to introduce new information and  effectively address data scarcity. Hence, Re-imagen~\citep{chen2022re,sheynin2022knn,li2022memory}  retrieval relevant training data from external databases to augment training data. 
\section{ Linear Extrapolation for Text-to-image Generation}
In this section, we begin by collecting similar images from the internet. Next, we explain how to extrapolate text descriptions. Following that, we use the extrapolated text-image pairs to train a diffusion model with RAT blocks. Finally, we sample images using NULL-condition guidance.
\subsection{Collecting Similar and Clean Images}
Linear extrapolation requires the images to be sufficiently close in semantic space. Hence, we automatically retrieve similar images by searching for their classification labels. However, search engines return both similar images and outliers. To eliminate unwanted outliers, we employ a cluster detector for irrelevant outliers and a classification detector for similar outliers. For the cluster detector, each image is encoded into a vector using the CLIP image encoder. Images retrieved with the same keyword are then clustered using K-means. If the distance from the cluster center to dataset images exceeds a threshold, this cluster is excluded. For the classification detector, we train a fine-grained classification model on the original dataset, which assigns  a label to each web image. If the label does not match with the search keyword, corresponding image is then excluded.

\subsection{Linear Extrapolation on Text Feature Space}
Here we introduce how to extrapolates text descriptions for web images. Assuming that web images are sufficiently close to dataset images in semantic space, each web image can be represented by nearest k images:
\begin{equation}
\argmin_W	|\textbf{f }-   \textbf{F} \times \textbf{w}|^2, 
\end{equation}
where $\textbf{w}=[w_1,w_2,...,w_k]$are the reconstruction weights and $\textbf{F} = [\textbf{f}_1,\textbf{f}_2,...,\textbf{f}_k]$ are the image features of dataset images produced by CLIP image encoder. Since the above equation is a super-determined problem, we solve this coefficient using least squares:

\begin{equation}
\textbf{w}=(\textbf{F}^T\textbf{F})^{-1}\textbf{F}^T \textbf{f}.
\end{equation}

We assume that the image feature space and text feature space share the same local manifold. Hence, the image reconstruction efficient $\textbf{w}$ can be used to compute the text feature of web images:

\begin{equation}
\textbf{s} = \textbf{S} \times \textbf{w},  
\end{equation}
 where $\textbf{S} = [\textbf{s}_1,\textbf{s}_2,...,\textbf{s}_k]$ is the fake sentence features for nearest k dataset images, and $\textbf{s}$ is the sentence feature for a web image.

\subsection{Recurrent Diffusion Transformer on Latent Space}
\begin{figure*}[t]
	\centering
	\includegraphics[width = \textwidth]{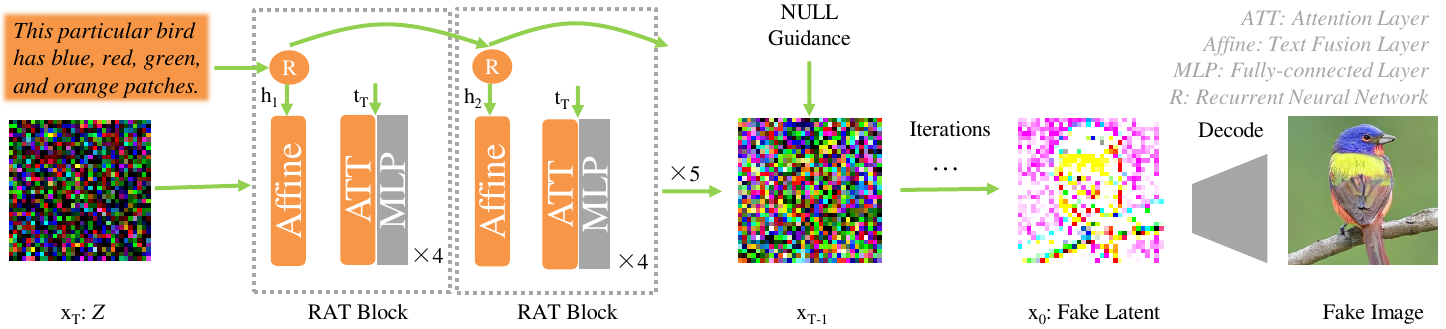}
	\caption{\label{Fig:rat} Latent diffusion model with recurrent affine transformation and NULL-guidance for text-to-image synthesis. The RAT blocks are connected by a recurrent neural network to ensure the global assignment of text information.}
	\label{Fig:framework}
\end{figure*}
The training objective of the diffusion model is the squared error loss proposed by DDPM~\citep{ho2020denoising}: 

\begin{equation}
L(\theta)=\left\|\epsilon-\epsilon_\theta\left(\sqrt{\overline{\bar{\alpha}}_t} \mathrm{x}_0+\sqrt{1-\bar{\alpha}_t} \epsilon \right)\right\|^2,
\end{equation}
where $\epsilon\in N(0,1)$ is the score noise injected at every diffusion step, and $\epsilon_{\theta}$ is the predicted noise by a diffusion network consisted of 12 transformer layers. $\overline{\bar{\alpha}}_t$ and $\bar{\alpha}_t$ are hyper-parameters controlling the speed of diffusion. The work of score mismatching~\citep{ye2024score} shows that predicting the score noise leads to an unbiased estimation.

\paragraph{Network architecture.} As depicted in Fig~\ref{Fig:rat}, the diffusion network consists of transformer blocks. Recurrent affine transformation is used to enhance the consistency between transformer blocks. To avoid directly mixing text embedding and time embedding, we stack four transformer blocks as a RAT block and text embedding is fed into  the top of each RAT block. 
Each RAT block  applies a channel-wise shifting operation on a image feature map:
\begin{equation}
	c'= c+\beta,
\end{equation}
where $c$ is the image feature vector  and $\beta$ is shifting parameters predicted by a one-hidden-layer multi-layer perception (MLP) conditioned on recurrent neural network hidden state $h_t$.

In each transformer block, we inject time embedding by a channel-wise scaling operation and a channel-wise shifting operation on $c$. At last, the image feature $c$ is multiplied by a  scaling parameter $\alpha$. This process can be formally expressed as:
\begin{equation}
	c'= Transformer((1+\gamma) \cdot c+\beta) \cdot \alpha,
\end{equation}

where $\alpha,\gamma,\beta$ are parameters predicted by two one-hidden-layer MLPs conditioned on time embedding.

When applied to an image feature map composed of $w\times h$ feature vectors, the same affine transformation is repeated for every feature vector. 
\paragraph{Early stop of fine-tuning.}   Extrapolation may produces training data very close to the original dataset, which makes fine-tuning saturate very quickly.
 Excessive fine-tuning epochs would forget knowledge gained from the extrapolated data and overfit  small datasets. As a result, the training loss of the diffusion model becomes unreliable. Therefore, fine-tuning should be stopped when the FID score begins to increase.

\subsection{Synthesizing Fake Images}
Finally, we introduces how to synthesizing images from scratch. As depicted in Figure~\ref{Fig:framework}, the synthesis begins with sampling a random vector $z$ from standard Gaussian distribution. And then, this noise is gradually denoised into an image latent code by the diffusion model. The reverse diffusion iterations are formulated as:
\begin{equation}
    \mathbf{x}_{t-1}=\frac{1}{\sqrt{\alpha_t}}\left(\mathbf{x}_t-\frac{1-\alpha_t}{\sqrt{1-\bar{\alpha}_t}} \epsilon_\theta\left(\mathbf{x}_t, t\right)\right)+\sigma_t \mathbf{z},
\end{equation}
where, $\alpha_t$, $\bar{\alpha}_t$ and $\sigma_t$ are diffusion hyper-parameters, and $z$ is a random vector sampled from standard Gaussian distribution. At last, we decode image latent  codes into images with the pre-trained decoder from Stable Diffusion~\citep{rombach2022high}.

\paragraph{NULL guidance.} A sentence with no new information is able to boost text-to-image performance obviously. This guidance is inspired by Classifier-free diffusion guidance~\citep{ho2022classifier} which uses a dummy class label to boost label-to-image performance.  Similarly, we design CLIP prompt without obvious visual meaning and embed them into the diffusion model. Specifically, we denote the original score estimation based on text description as $\epsilon_{text}$ and score estimation based on null description as $\epsilon_{null}$. Then we mix these two estimations for a more accurate estimation $\epsilon'$:
\begin{equation}
    \epsilon'=(\epsilon_{text}-\epsilon_{null}) \times \eta + \epsilon_{null},
\end{equation}
where, $\eta$ is the guidance ration controlling the balance of two estimations. When $\eta=1$, NULL Guidance falls back to an ordinary score estimation. Usually, a NULL prompt with the average meaning of the dataset achieve the best performance.  
\section{Experiments}
	\begin{figure*}[t!h]
		\centering
		
		\begin{minipage}[c]{0.01\textwidth}
			\fontsize{2.0pt}{0.5\baselineskip}\selectfont \center{\ } 
		\end{minipage}
		\hfill
		\begin{minipage}[t]{0.115\textwidth}
			\center{\scriptsize{A small bird with blue-grey wings, rust colored sides and white collar.}}
		\end{minipage}
		\hfill
		\begin{minipage}[t]{0.115\textwidth}
			\center{\scriptsize{This bird is white from crown to belly, with gray wingbars and retrices.}} 
		\end{minipage}
		\hfill
		\begin{minipage}[t]{0.115\textwidth}
			\center{\scriptsize{This bird is mainly grey, it has brown on the feathers and back of the tail.}}
		\end{minipage}
		\hfill
		\begin{minipage}[t]{0.115\textwidth}
			\center{\scriptsize{A bird with blue head, white belly and breast, and the bill is pointed.}}
		\end{minipage}
		\hspace{1pt}
		\begin{minipage}[t]{0.115\textwidth}
			\center{\scriptsize{This flower has a lot of dark red petals and no visible outer stigma or stamen.}} 
		\end{minipage}
		\hfill
		\begin{minipage}[t]{0.115\textwidth}
			\center{\scriptsize{This flower has petals that are purple and bunched together.}} 
		\end{minipage}
		\hfill
		\begin{minipage}[t]{0.115\textwidth}
			\center{\scriptsize{This flower has large smooth white petals that turn yellow toward the center.}}
		\end{minipage}
		\hfill
		\begin{minipage}[t]{0.115\textwidth}
			\center{\scriptsize{A pale purple five petaled flower with yellow stamen and green stigma.}}
		\end{minipage}
		\vspace{2pt}
		
		\begin{minipage}[c]{0.02\textwidth}
			\center{\rotatebox{90}{GT}}
		\end{minipage}
		\hfill
		\begin{minipage}{0.115\textwidth}
			\includegraphics[width=\textwidth]{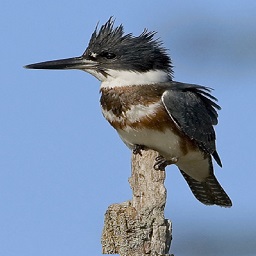}
		\end{minipage}
		\hfill
		\begin{minipage}{0.115\textwidth}
			\includegraphics[width=\textwidth]{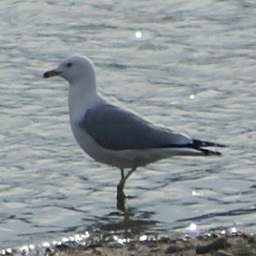}
		\end{minipage}
		\hfill
		\begin{minipage}{0.115\textwidth}
			\includegraphics[width=\textwidth]{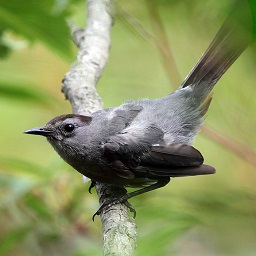}
		\end{minipage}
		\hfill
		\begin{minipage}{0.115\textwidth}
			\includegraphics[width=\textwidth]{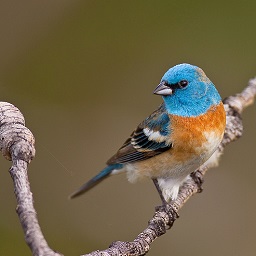}
		\end{minipage}
		\hspace{1pt}
		\begin{minipage}{0.115\textwidth}
			\includegraphics[width=\textwidth]{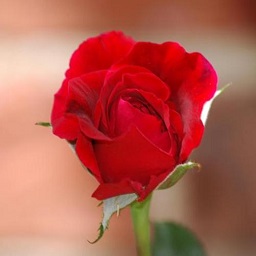}
		\end{minipage}
		\hfill
		\begin{minipage}{0.115\textwidth}
			\includegraphics[width=\textwidth]{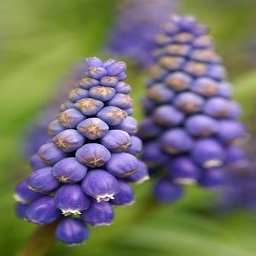}
		\end{minipage}
		\hfill
		\begin{minipage}{0.115\textwidth}
			\includegraphics[width=\textwidth]{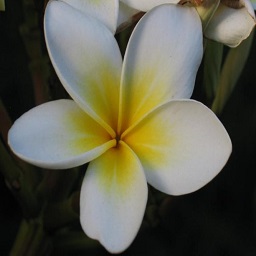}
		\end{minipage}
		\hfill
		\begin{minipage}{0.115\textwidth}
			\includegraphics[width=\textwidth]{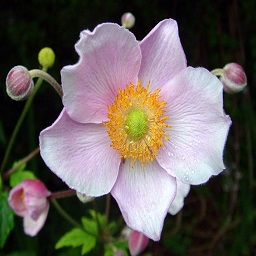}
		\end{minipage}
		\vspace{2pt}

		\begin{minipage}[c]{0.02\textwidth}
			\center{\rotatebox{90}{DF-GAN}}
		\end{minipage}
		\hfill
		\begin{minipage}{0.115\textwidth}
			\includegraphics[width=\textwidth]{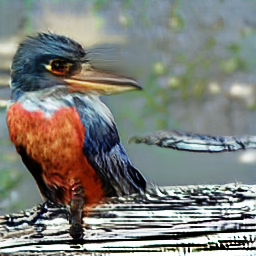}
		\end{minipage}
		\hfill
		\begin{minipage}{0.115\textwidth}
			\includegraphics[width=\textwidth]{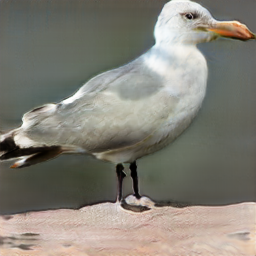}
		\end{minipage}
		\hfill
		\begin{minipage}{0.115\textwidth}
			\includegraphics[width=\textwidth]{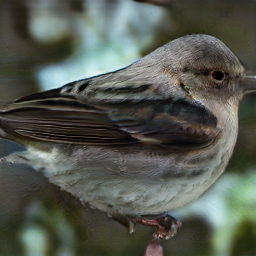}
		\end{minipage}
		\hfill
		\begin{minipage}{0.115\textwidth}
			\includegraphics[width=\textwidth]{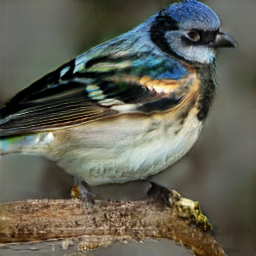}
		\end{minipage}
		\hspace{1pt}
		\begin{minipage}{0.115\textwidth}
			\includegraphics[width=\textwidth]{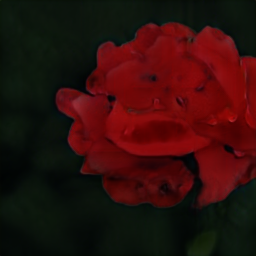}
		\end{minipage}
		\hfill
		\begin{minipage}{0.115\textwidth}
			\includegraphics[width=\textwidth]{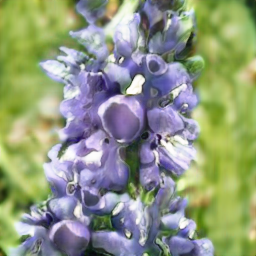}
		\end{minipage}
		\hfill
		\begin{minipage}{0.115\textwidth}
			\includegraphics[width=\textwidth]{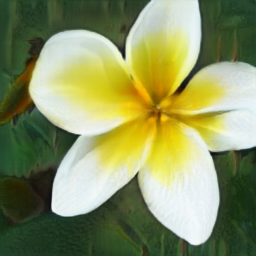}
		\end{minipage}
		\hfill
		\begin{minipage}{0.115\textwidth}
			\includegraphics[width=\textwidth]{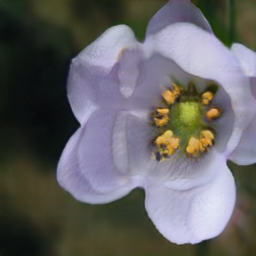}
		\end{minipage}
		\vspace{2pt}
		
		\begin{minipage}[c]{0.02\textwidth}
			\center{\rotatebox{90}{RAT-GAN}}
		\end{minipage}
		\hfill
		\begin{minipage}{0.115\textwidth}
			\includegraphics[width=\textwidth]{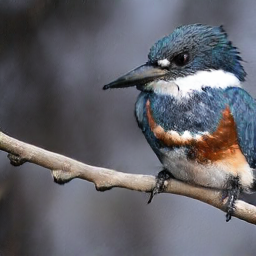}
		\end{minipage}
		\hfill
		\begin{minipage}{0.115\textwidth}
			\includegraphics[width=\textwidth]{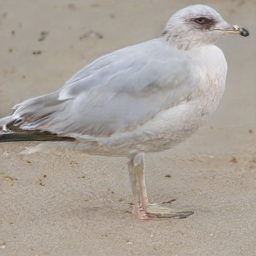}
		\end{minipage}
		\hfill
		\begin{minipage}{0.115\textwidth}
			\includegraphics[width=\textwidth]{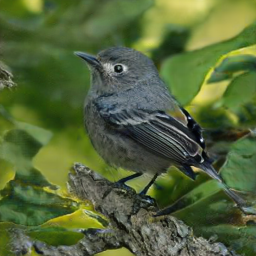}
		\end{minipage}
		\hfill
		\begin{minipage}{0.115\textwidth}
			\includegraphics[width=\textwidth]{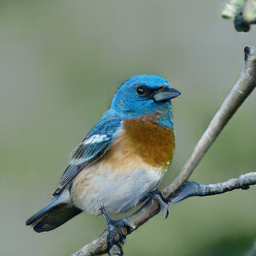}
		\end{minipage}
		\hspace{1pt}
		\begin{minipage}{0.115\textwidth}
			\includegraphics[width=\textwidth]{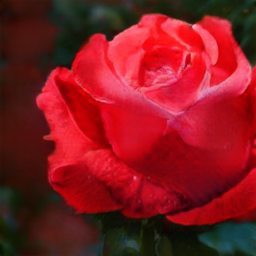}
		\end{minipage}
		\hfill
		\begin{minipage}{0.115\textwidth}
			\includegraphics[width=\textwidth]{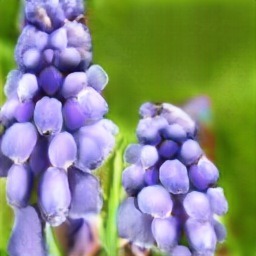}
		\end{minipage}
		\hfill
		\begin{minipage}{0.115\textwidth}
			\includegraphics[width=\textwidth]{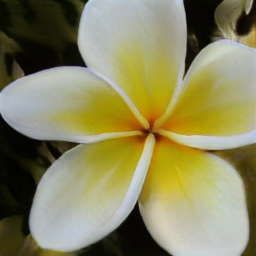}
		\end{minipage}
		\hfill
		\begin{minipage}{0.115\textwidth}
			\includegraphics[width=\textwidth]{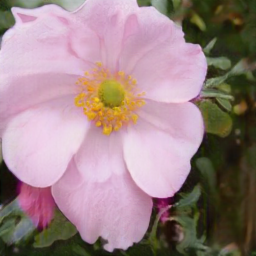}
		\end{minipage}
    \vspace{2pt}
    
  \begin{minipage}[c]{0.02\textwidth}
			\center{\rotatebox{90}{Ours}}
		\end{minipage}
		\hfill
		\begin{minipage}{0.115\textwidth}
			\includegraphics[width=\textwidth]{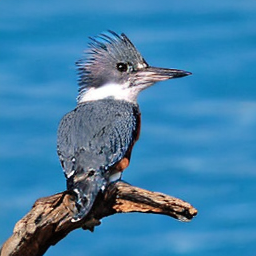}
		\end{minipage}
		\hfill
		\begin{minipage}{0.115\textwidth}
			\includegraphics[width=\textwidth]{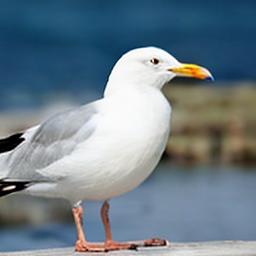}
		\end{minipage}
		\hfill
		\begin{minipage}{0.115\textwidth}
			\includegraphics[width=\textwidth]{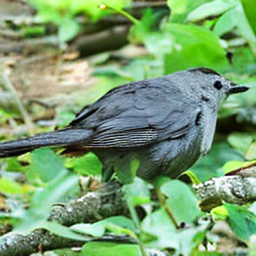}
		\end{minipage}
		\hfill
		\begin{minipage}{0.115\textwidth}
			\includegraphics[width=\textwidth]{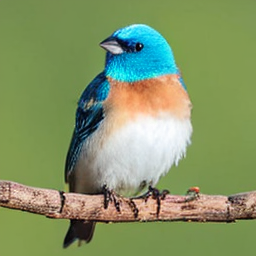}
		\end{minipage}
		\hspace{1pt}
		\begin{minipage}{0.115\textwidth}
			\includegraphics[width=\textwidth]{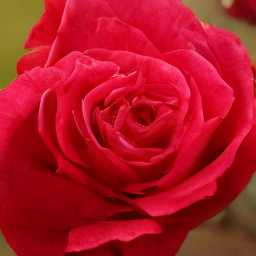}
		\end{minipage}
		\hfill
		\begin{minipage}{0.115\textwidth}
			\includegraphics[width=\textwidth]{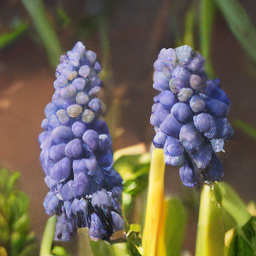}
		\end{minipage}
		\hfill
		\begin{minipage}{0.115\textwidth}
			\includegraphics[width=\textwidth]{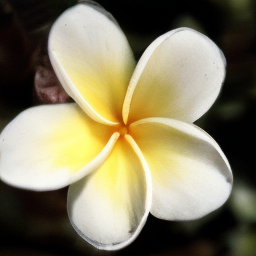}
		\end{minipage}
		\hfill
		\begin{minipage}{0.115\textwidth}
			\includegraphics[width=\textwidth]{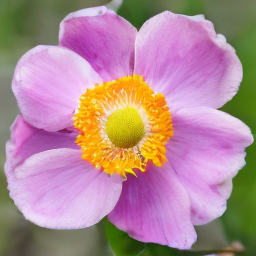}
		\end{minipage}
		\vspace{2pt}
		\vspace{2pt}
		\caption{Qualitative comparison on the CUB and Oxford dataset. The input text descriptions are given in the first row and the corresponding generated images from different methods are shown in the same column. Best view in color and zoom in.}
		\label{fig:qualitative_cub}
		\vspace{-4mm}
	\end{figure*}

\paragraph{Datasets.} We report results on the popular CUB, Oxford-102, and MS COCO datasets. The CUB dataset includes 200 categories with a total of 11,788 bird images, while the Oxford-102 dataset contains 102 categories with 8,189 flower images. Unlike the approaches taken in \cite{reed2016learning,reed2016generative}, we utilize the entire dataset for both training and testing. Each image is paired with 10 captions. To expand the original datasets, we collect 300,000 bird images and 130,000 flower images. The MS COCO dataset comprises 123,287 images, each with 5 sentence annotations. We use the official training split of COCO for training and the official validation split for testing. During mini-batch selection, a random image view (e.g., crop or flip) is chosen for one of the captions.
\paragraph{Web images.} For the CUB and Oxford datasets, we collected 603,484 bird images and 331,602 flower images using search engines, utilizing fine-grained classification labels as search keywords. After removing detected outliers, we retained 399,246 bird images and 132,648 flower images. In the case of the COCO dataset, we gathered 770,059 daily images without applying any outlier detection, as the precise descriptions in COCO allow search engines to retrieve clean images effectively.

	\paragraph{Training details.}
	The text encoder is a pre-trained CLIP text encoder with an output of size $512$. The latent encoder and decoder is pre-trained by Stable Diffusion~\citep{rombach2022high}. We have tried to pre-train new latent encoders on extrapolated data but the results are not satisfying.
	Adam optimizer is used to optimize the network with base learning rates of 	0.0001 and weight decay of 0. The same as RAT-GAN, we used a mini-batch size of 24 to train the model. Most training and testing of our model are conducted on 2 RTX 3090 Ti and the detailed training consumption is listed in Table~\ref{tab:consum}.
 \paragraph{Evaluation metrics.}

	We adopt the widely used Inception Score (IS)~\citep{DBLP:conf/nips/SalimansGZCRCC16} and Fr\'{e}chet Inception Distance (FID)~\citep{DBLP:conf/nips/HeuselRUNH17} to quantify the performance. On the MS COCO dataset, an Inception-v3 network pre-trained on the ImageNet dataset is used to compute the KL-divergence between the conditional class distribution (generated images) and the marginal class distribution (real images). The presence of a large IS indicates that the generated images are of high quality. The FID computes the Fr\'{e}chet Distance between the image feature distributions of the generated and real-world images. The image features are extracted by the same pre-trained Inception v3 network. A lower FID implies the generated images are closer to the real images.   We only compare the FID on the COCO dataset. On the CUB and Oxford-102 dataset, pre-trained Inception models are fine-tuned on two fine-grained classification tasks~\citep{DBLP:journals/pami/ZhangXLZWHM19}. 

There are two conflicts in evaluation methods in previous works. First, some studies report Inception Score (IS) using the ImageNet Inception model, while others use a fine-tuned version. Second, some works evaluate using the entire training data, whereas others use only the test split. To address these inconsistencies, we report IS and FID using both Inception models and employ the same Inception model as DM-GAN for consistency. Additionally, to resolve conflicts related to data splits, we report the FID scores of our model and other re-implemented models using the entire dataset for both training and testing. According to results from RAT-GAN~\citep{ye2023recurrent}, training and testing on the full dataset typically yields the best FID scores. We will also release all evaluation codes on GitHub.
    
	\paragraph{Compared models.} We compare our model with recent state-of-the-art methods: StackGAN++~\citep{DBLP:journals/pami/ZhangXLZWHM19}, DM-GAN~\citep{DBLP:conf/cvpr/ZhuP0019}, DF-GAN~\citep{DBLP:journals/corr/abs-2008-05865}, DAE-GAN~\citep{DBLP:journals/corr/abs-2108-12141}, VQ-diffusion~\citep{gu2022vector}, AttnGAN~\citep{DBLP:conf/cvpr/XuZHZGH018},
	GALIP~\citep{DBLP:conf/ijcnn/ZhangS21},U-ViT~\citep{bao2023all}, and RAT-GAN~\citep{ye2023recurrent}. 
	\subsection{Comparisons with Others}
	\begin{table*}[t!]
		\caption{Performance of IS and FID of StackGAN++, AttnGAN, SSGAN, DM-GAN, DTGAN, DF-GAN and our method on the CUB, Oxford and MS COCO datasets. The results are taken from the authors' own papers.
			The best results are in bold.}
		\label{tab:results}
		\vspace{-2mm}
		\centering
		\begin{adjustbox}{max width=1\textwidth}
			\begin{tabular}{lccccccccc}
				\toprule
				\multirow{2}*{Methods} &\multicolumn{2}{c}{IS(Fine-tune) $\uparrow$}&\multicolumn{2}{c}{IS(ImageNet) $\uparrow$}& \multicolumn{2}{c}{FID(Fine-tune) $\downarrow$}&\multicolumn{3}{c}{FID(ImageNet) $\downarrow$} \cr 
				\cmidrule(lr){2-3} \cmidrule(lr){4-5}\cmidrule(lr){6-7} \cmidrule(lr){8-10}
				&CUB&Oxford&CUB&Oxford&CUB& Oxford&CUB&Oxford& COCO\\
				\midrule 
				StackGAN++   & 4.04  &3.26& 4.04   &3.26  &23.96&48.68&15.30&32.33 &81.59\\
				AttnGAN            & 4.36 &-  & 4.36&-   &-  &-&23.98&-&35.49\\

				DAE-GAN                       & 4.42  &-& - &-  & -&-&15.19&- & 28.12 \\
				DM-GAN                 & 4.75  &-&- &-  &- &-&16.09&- &32.64\\
				DF-GAN              & 5.10   &3.80 & 4.96  &3.92  &17.23 &18.90&14.81& 22.56&21.42\\

				RAT-GAN                                   & {5.36 }&{4.09}& {5.00 }&{3.95}   &{13.91} &16.04&10.21&18.68& {14.60}\\
                GALIP            & -  &-  & -  &-  &- &-&10.05&-&{5.85}\\
                VQ-Diffusion                                  & -&-& -&-& -&-&10.32&14.10 & {13.86}\\
                U-ViT            & -  &-  & -  &-  &- &-&-&-&{5.45}\\
    Ours                                   & \textbf{6.56 }&\textbf{4.35} & \textbf{6.37 }&\textbf{4.11}   &\textbf{7.91} &\textbf{8.58}&\textbf{6.36}&\textbf{9.52}& \textbf{5.00}\\
				\bottomrule
			\end{tabular}
		\end{adjustbox}
		\vspace{-2mm}
	\end{table*}
 \paragraph{Quantitative results.} We present results for the CUB dataset of bird images, the Oxford-102 dataset of flower images, and the MS COCO dataset of common objects, as shown in Table~\ref{tab:results}. On the CUB dataset, our model achieve an IS score of 6.56 and an FID score of 6.36, outperforming all the previous models. For the Oxford dataset, we achieve an IS score of 4.35 and an FID score of 6.36, outperforming all the previous models.   On the COCO dataset, our model achieves an FID score of 5.00 that is competitive with previous best result.Compared with VQ-Diffusion, our model uses less training data and achieve much better performance. This comparison reveals that pre-training on large datasets can be inefficient and lead to suboptimal results. 
 Moreover, results in Table~\ref{tab:results} reveal that Inception model pre-trained on ImageNet is less sensitive than fine-tuned on small datasets.
 Additionally, the Inception score on the Oxford dataset exceeds that of real images (4.10). Extensive results demonstrate the effectiveness and generalization ability of the proposed data extrapolation method.

\paragraph {Qualitative results.}
We present qualitative results for the CUB dataset of bird images and the Oxford-102 dataset of flower images. In Figure~\ref{fig:qualitative_cub}
, we compare the visualization results of DF-GAN, RAT-GAN, and our model. DF-GAN and RAT-GAN are previous state-of-the-art methods for text-to-image synthesis.
On the CUB dataset, with more clear details such as feathers, eyes, and feet, our model clearly outperforms DF-GAN and RAT-GAN. Additionally, the background in our model's results is more coherent compared to RAT-GAN. On the Oxford dataset, our model exhibits better texture and more relevant colors than the others. With the proposed text extrapolation, RAT block, and null-guidance, our model demonstrates fewer distorted shapes and more relevant content compared to the other two models.

\begin{figure}[t!h]
	\centering
	
	\begin{minipage}[c]{0.01\textwidth}
		\fontsize{2.0pt}{0.5\baselineskip}\selectfont \center{\ } 
	\end{minipage}
	\hfill
	\begin{minipage}[t]{0.18\textwidth}
		\center{\scriptsize{A police man on a motorcycle is idle in front of a bush.}}
	\end{minipage}
	\hfill
	\begin{minipage}[t]{0.18\textwidth}
		\center{\scriptsize{A man riding a wave on top of a surfboard.}} 
	\end{minipage}
	\hfill
	\begin{minipage}[t]{0.18\textwidth}
		\center{\scriptsize{Some red and\\ green flower in a room.}}
	\end{minipage}
	\hfill
	\begin{minipage}[t]{0.18\textwidth}
		\center{\scriptsize{Assorted electronic devices sitting together in a photo.}}
	\end{minipage}
 \begin{minipage}[t]{0.18\textwidth}
		\center{\scriptsize{An elephant raising its truck with a tree in background.}}
	\end{minipage}
	
	\vspace{3pt}

	
	
	
	\begin{minipage}[c]{0.01\textwidth}
		\center{\rotatebox{90}{RAT-GAN}}
	\end{minipage}
	\hfill
	\begin{minipage}{0.18\textwidth}
		\includegraphics[width=\textwidth]{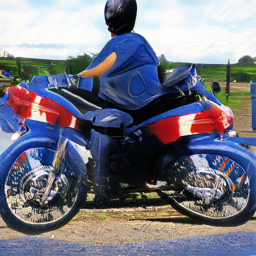}
	\end{minipage}
	\hfill
	\begin{minipage}{0.18\textwidth}
		\includegraphics[width=\textwidth]{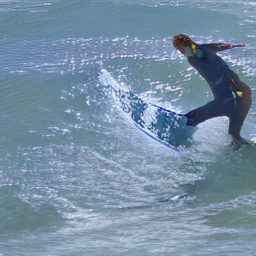}
	\end{minipage}
	\hfill
	\begin{minipage}{0.18\textwidth}
		\includegraphics[width=\textwidth]{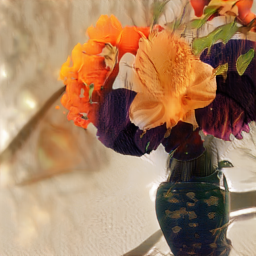}
	\end{minipage}
	\hfill
	\begin{minipage}{0.18\textwidth}
		\includegraphics[width=\textwidth]{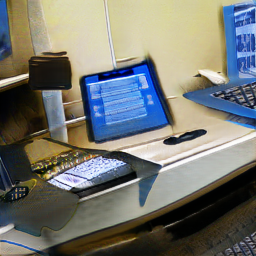}
	\end{minipage}	
 \hfill
	\begin{minipage}{0.18\textwidth}
		\includegraphics[width=\textwidth]{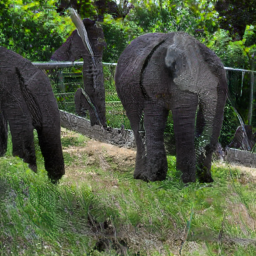}
	\end{minipage}	
	
	\vspace{3pt}
	
	\begin{minipage}[c]{0.01\textwidth}
		\center{\rotatebox{90}{Ours}}
	\end{minipage}
	\hfill
	\begin{minipage}{0.18\textwidth}
		\includegraphics[width=\textwidth]{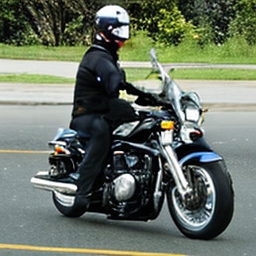}
	\end{minipage}
	\hfill
	\begin{minipage}{0.18\textwidth}
		\includegraphics[width=\textwidth]{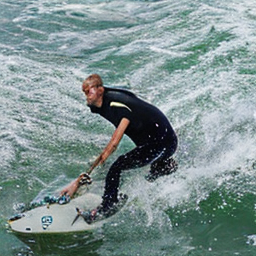}
	\end{minipage}
	\hfill
	\begin{minipage}{0.18\textwidth}
		\includegraphics[width=\textwidth]{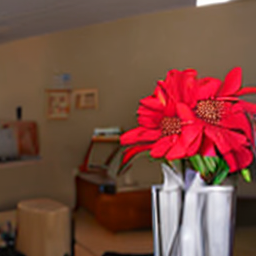}
	\end{minipage}
	\hfill
	\begin{minipage}{0.18\textwidth}
		\includegraphics[width=\textwidth]{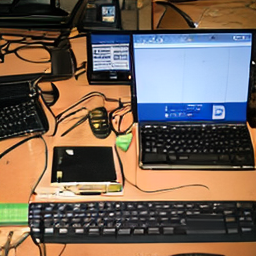}
	\end{minipage}
 \hfill
	\begin{minipage}{0.18\textwidth}
		\includegraphics[width=\textwidth]{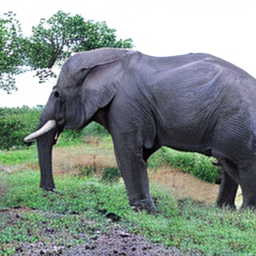}
	\end{minipage}

	\caption{Qualitative comparison of our model with RAT-GAN on the COCO dataset. }
	\label{Fig:coco}
	\vspace{-4mm}
\end{figure}
The qualitative results for the COCO dataset are shown in Figure~\ref{Fig:coco}. The COCO dataset includes a wide variety of common objects, which makes it particularly susceptible to the long-tail problem~\citep{chen2022re}. With additional training data obtained through extrapolation, our model generates more realistic objects compared to RAT-GAN. However, the collected 770,059 images are still insufficient to cover the entire distribution of images in COCO. As a result, the outputs from COCO are not as realistic as those from the CUB and Oxford datasets.

	\subsection{Ablation Studies}
	\begin{table*}[t!]
		\caption{Ablation studies on the CUB dataset. We utilize a NULL-guidance ratio of 1.5 during sampling. The FID score was employed to evaluate generation performance.}
		\label{tab:ablation_study_components}
		\vspace{-4mm}
		\begin{center}
			\begin{tabular}{c c c c c c c c c c c}
				\hline
				\multirow{2}{*}{ID}& \multicolumn{4}{c}{Component}& \multicolumn{6}{c}{Extrapolation Quantity(k)} \\

                \cline{2-5}&Cluster&Classification  &RAT& NULL& 0& 50&100&200&300&400\\ \cline{6-10}
				\hline 
				0 & - & - & - & -  & 16.74& - & - & -& -  & 30.78 \\
				\hline
    1 & \checkmark & - & - & -  & -& - & - & -& -  & 20.67 \\
				\hline
				2 & - & \checkmark& -  & -  & -& - & - & -& -  & 12.45 \\
    \hline
    3 & \checkmark & \checkmark& -  & -  & -& - & - & -& -  & 9.87 \\
				\hline
				4 & \checkmark  & \checkmark& \checkmark & -& -  & -& - & - & -  & 8.76\\
				\hline   
				5 & \checkmark & \checkmark & - &\checkmark & -& -& - & - & -  & 7.65\\                     
				\hline
				6 &\checkmark & \checkmark&\checkmark  &\checkmark & -&  9.56& 7.34 & 6.87  & 6.54 & \textbf{6.36}\\
				\hline
			\end{tabular}
		\end{center}
		\vspace{-4mm}
	\end{table*}

\paragraph{Analysis of outlier detectors.} In Table~\ref{tab:ablation_study_components}, we present text-to-image results without cluster detector or classification detector. According to ID 0,1 and 2, the FID score without outlier detectors degrade severely because noisy images force the diffusion model to generate irrelevant objects.  Although fine-tuning on small datasets could alleviate noise pollution but parameters also forget general knowledge at the same time. According to ID 2 and 3, classification detector performs better than cluster detector because it has utilized fine-grained classification labels. 
    \paragraph{Analysis of extrapolation quantity.}More images generally lead to improved text-to-image results. however, this trend saturates around 100,000 images, after which the improvement in FID becomes less significant with more training samples. This phenomenon aligns with that diffusion models perform much better than GANs on the COCO dataset (84K images) but exhibit similar performance to GANs on the CUB and Oxford datasets(~10K images). Furthermore, with transformers as core building blocks,  GALIP performs similarly to previous models on the CUB dataset. This suggests that transformer architectures exacerbate the need for larger training datasets.
	\begin{table}[t!]
     \caption{\label{tab:consum} Training consumption on the CUB, Oxford and COCO datasets. Fine-tuning is performed on the original dataset until the FID scores increase.  }
    \centering
    \begin{tabular}{l |l llll}

			\hline
			  {Dataset}&Device&  Original dataset&Extrapolated data& Fine-tuning\\ 
			\hline 
   
             	CUB&2  RTX 3090 Ti&5 days/1500 epochs&10 days/100 epochs&6 hours/50 epochs\\ 
			 Oxford&2  RTX 3090 Ti&5 days/1500 epochs&8 days/200 epochs&6 hours/50 epochs\\
            COCO&2  RTX 4090&10 days/125 epochs&20 days/95 epochs&7 days/70 epochs\\ 
            \hline

		\end{tabular}
\end{table}

     \begin{wraptable}{r}{0.55\textwidth}
     \caption{\label{tab:null}The impact of various NULL prompts on FID scores in the CUB dataset.  }
    \centering
    \begin{tabular}{c |c cccc}

			\hline
			  \multirow{2}{*}{NULL Prompts}& \multicolumn{3}{c}{Guidance Ratio} \\
                & 1.25&1.5&2.0\\ 
			\hline 
   
             	``Null"&7.23&7.16&7.68\\ 
			``a picture" &6.89&6.54&7.14\\
            ``no description"&6.97&6.47&7.25\\ 
			``a picture of bird"&6.46&6.36&6.86 \\
			``a picture of flower" &9.04&10.6&11.4\\
            ``we don't know what it is"&8.98&9.35&9.94 \\

		\end{tabular}

\end{wraptable}

	\paragraph{Analysis of NULL guidance.}The performance of NULL guidance is influenced by both the NULL prompt and the guidance ratio. The results in Table~\ref{tab:null} indicate that a NULL prompt reflecting the average meaning of the dataset achieves the best performance. Additionally, a suitable guidance ratio is crucial for optimal results, and we find that a ratio around 1.5 yields the best performance on the CUB and COCO datasets. However, on the Oxford dataset, NULL guidance improves the Inception Score from 4.10 to 4.35 but degrades the FID score from 9.52 to 11.07.

\begin{wraptable}{}{0.55\textwidth}
     \caption{\label{tab:coco_aba} Ablation studies on the MS COCO dataset.  We adopt ``A picture'' as the NULL prompt.}
    \centering
    \begin{tabular}{c |c cccc}

			\hline
			  \multirow{2}{*}{Training data}& \multicolumn{3}{c}{FID score} \\
                & $\eta = 1.0$&$\eta = 1.5$&$\eta = 2.0$\\ 
			\hline 
   
             	COCO&11.89&7.99&8.43\\ 
			 Extrapolation&12.33&8.41&9.24\\
            COCO-ft&8.45&5.00&5.56\\ 
			\hline

		\end{tabular}

\end{wraptable} 
    \paragraph{Analysis of text injection.} Text injection is crucial for text-to-image generation. As shown in ID 4 and 5 of Table~\ref{tab:ablation_study_components}, RAT significantly improves the FID score. Further experiments indicate that directly mixing text feature with time embedding results in an FID score of 25.41, which is much worse than 16.74 achieved by RAT. This suggests that time embedding provides information very different to text embedding. Additionally, incorporating a scaling operator into RAT can lead to model collapse, as information becomes highly compressed in latent space. Consequently, the mean value of the latent code becomes sensitive, and the scaling operation disrupts the information structure.
    
    \paragraph{Ablation studies on the MS COCO dataset.} 
We conduct ablation studies on the MS COCO dataset, as presented in Table~\ref{tab:coco_aba}. The MS COCO dataset differs significantly from the CUB and Oxford datasets in terms of variety and image quantity. Experimental results demonstrate that linear extrapolation and fine-tuning (5.00) outperform the original COCO dataset (7.99). However, unlike CUB and Oxford, fine-tuning on COCO requires much more time, as shown in Table~\ref{tab:consum}. Additionally, we observe that early stopping is unnecessary for fine-tuning on the COCO dataset due to its larger image volume compared to CUB and Oxford.

In Table~\ref{tab:coco_para}, we compare the pre-training dataset and model parameters with previous models on the MS COCO dataset. The compared models are all pre-trained on external datasets and fine-tuned on MS COCO dataset. Our result outperforms all previous models except for Parti but we use much less pre-training images and parameters than Parti. Moreover, our diffusion model is designed for small datasets and requires very few GPUs for training.

\begin{figure}[]
	\centering

	\begin{minipage}{0.092\textwidth}
		\includegraphics[width=\textwidth]{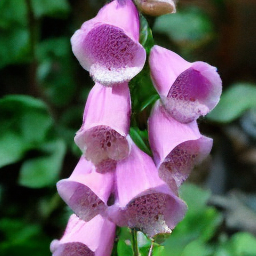}
	\end{minipage}
	\begin{minipage}{0.092\textwidth}
		\includegraphics[width=\textwidth]{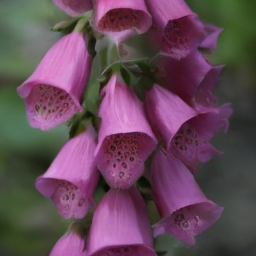}
	\end{minipage}
	\begin{minipage}{0.092\textwidth}
		\includegraphics[width=\textwidth]{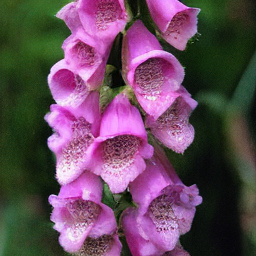}
	\end{minipage}
	\begin{minipage}{0.092\textwidth}
		\includegraphics[width=\textwidth]{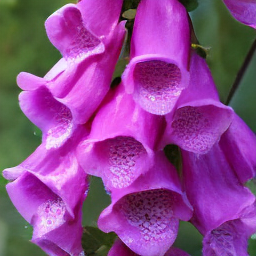}
	\end{minipage}
	\begin{minipage}{0.092\textwidth}
		\includegraphics[width=\textwidth]{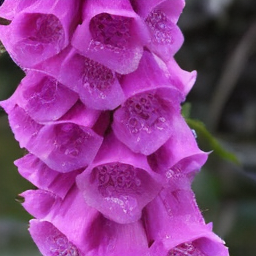}
	\end{minipage}

		\begin{minipage}{0.092\textwidth}
		\includegraphics[width=\textwidth]{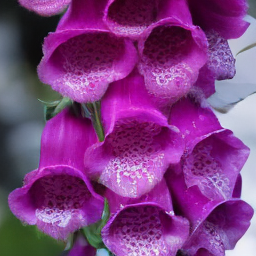}
	\end{minipage}
	\begin{minipage}{0.092\textwidth}
		\includegraphics[width=\textwidth]{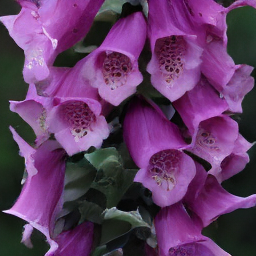}
	\end{minipage}
	\begin{minipage}{0.092\textwidth}
		\includegraphics[width=\textwidth]{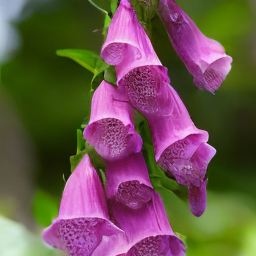}
	\end{minipage}
	\begin{minipage}{0.092\textwidth}
		\includegraphics[width=\textwidth]{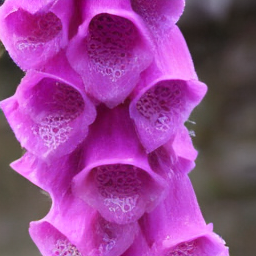}
	\end{minipage}
	\begin{minipage}{0.092\textwidth}
		\includegraphics[width=\textwidth]{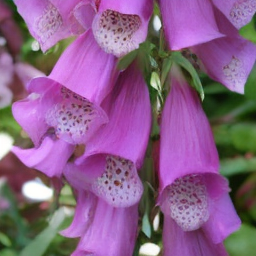}
	\end{minipage}
\begin{minipage}[t]{1\textwidth}
		\center{This flower is purple and white, and has petals that are bulb shaped and drooping downward.}
	\end{minipage}
	\caption{Randomly generated images from the Oxford dataset.  Best view in color and zoom in. }
	\label{Fig:variaty}
\end{figure}

    \paragraph{Diversity.} To qualitatively evaluate the diversity of our proposed model, we generate random images conditioned on the same text description and different random noises. In Figure~\ref{Fig:variaty}, we present 10 images generated from the same text. These images exhibit similar foreground elements while showcasing high diversity in spatial structure, demonstrating that our model effectively controls the image content.

\begin{table}[t]
    \centering
    \scalebox{0.85}{
    \begin{tabular}{lcccc}
    \toprule
        Model & FID & Type & Pre-training images & \#Params \\

\hline
         \quad Parti~\citep{yu2022scaling} & \textbf{3.22} & Autoregressive & 4.8B & 20B \\
         \quad Make-A-Scene~\citep{gafni2022make} & 7.55 & Autoregressive & 35M & 4B \\
         \quad Re-Imagen~\citep{chen2022re} & 5.25 & Diffusion & 50M & 2.5B  \\
         \quad VQ-Diffusion~\citep{gu2022vector} & 19.75 & Diffusion & 15M & 370M \\
\quad Ours & {5.00} & Diffusion & 7M & 464M\\
\hline
    \end{tabular}}
    \caption{Comparison of pre-training dataset and parameter quantity of different models on the MS COCO dataset. Parameters for text encoder, latent encoder and super resolution are not counted. }\label{tab:coco_para}
\end{table}

 \section{Conclusion and Future work}
In this paper, we propose a new data augmentation method for text-to-image generation using linear extrapolation. Specifically, we apply linear extrapolation only on text data, and new image data are retrieved from the internet by search engines. For the reliability of new text-image pairs, we design two outlier detectors to purify retrieved images. Based on extrapolation, we construct training samples dozens of times larger than the original dataset, resulting in a significant improvement in text-to-image performance. Moreover, we propose a NULL-condition guidance to refine the score estimation for text-to-image generation. This guidance is also applicable to existing text-to-image models without further training. In the future, linear extrapolation and  NULL-condition guidance  could be applied to tasks beyond text-to-image generation.

\bibliography{iclr2025_conference}

\begin{thebibliography}{49}
\providecommand{\natexlab}[1]{#1}
\providecommand{\url}[1]{\texttt{#1}}
\expandafter\ifx\csname urlstyle\endcsname\relax
  \providecommand{\doi}[1]{doi: #1}\else
  \providecommand{\doi}{doi: \begingroup \urlstyle{rm}\Url}\fi

\bibitem[Amodei et~al.(2016)Amodei, Ananthanarayanan, Anubhai, Bai, Battenberg, Case, Casper, Catanzaro, Cheng, Chen, et~al.]{amodei2016deep}
Dario Amodei, Sundaram Ananthanarayanan, Rishita Anubhai, Jingliang Bai, Eric Battenberg, Carl Case, Jared Casper, Bryan Catanzaro, Qiang Cheng, Guoliang Chen, et~al.
\newblock Deep speech 2: End-to-end speech recognition in english and mandarin.
\newblock In \emph{International conference on machine learning}, pp.\  173--182. PMLR, 2016.

\bibitem[Bao et~al.(2023)Bao, Nie, Xue, Cao, Li, Su, and Zhu]{bao2023all}
Fan Bao, Shen Nie, Kaiwen Xue, Yue Cao, Chongxuan Li, Hang Su, and Jun Zhu.
\newblock All are worth words: A vit backbone for diffusion models.
\newblock In \emph{Proceedings of the IEEE/CVF conference on computer vision and pattern recognition}, pp.\  22669--22679, 2023.

\bibitem[Brack et~al.(2024)Brack, Friedrich, Kornmeier, Tsaban, Schramowski, Kersting, and Passos]{brack2024ledits++}
Manuel Brack, Felix Friedrich, Katharia Kornmeier, Linoy Tsaban, Patrick Schramowski, Kristian Kersting, and Apolin{\'a}rio Passos.
\newblock Ledits++: Limitless image editing using text-to-image models.
\newblock In \emph{Proceedings of the IEEE/CVF Conference on Computer Vision and Pattern Recognition}, pp.\  8861--8870, 2024.

\bibitem[Chen et~al.(2022)Chen, Hu, Saharia, and Cohen]{chen2022re}
Wenhu Chen, Hexiang Hu, Chitwan Saharia, and William~W Cohen.
\newblock Re-imagen: Retrieval-augmented text-to-image generator.
\newblock \emph{arXiv preprint arXiv:2209.14491}, 2022.

\bibitem[Feng et~al.(2022)Feng, Niu, Li, and Wang]{DBLP:journals/tmm/FengNLW22}
Fangxiang Feng, Tianrui Niu, Ruifan Li, and Xiaojie Wang.
\newblock Modality disentangled discriminator for text-to-image synthesis.
\newblock \emph{{IEEE} Trans. Multim.}, 24:\penalty0 2112--2124, 2022.

\bibitem[Gafni et~al.(2022)Gafni, Polyak, Ashual, Sheynin, Parikh, and Taigman]{gafni2022make}
Oran Gafni, Adam Polyak, Oron Ashual, Shelly Sheynin, Devi Parikh, and Yaniv Taigman.
\newblock Make-a-scene: Scene-based text-to-image generation with human priors.
\newblock In \emph{European Conference on Computer Vision}, pp.\  89--106. Springer, 2022.

\bibitem[Graves et~al.(2013)Graves, Mohamed, and Hinton]{graves2013speech}
Alex Graves, Abdel-rahman Mohamed, and Geoffrey Hinton.
\newblock Speech recognition with deep recurrent neural networks.
\newblock In \emph{2013 IEEE international conference on acoustics, speech and signal processing}, pp.\  6645--6649. Ieee, 2013.

\bibitem[Gu et~al.(2022)Gu, Chen, Bao, Wen, Zhang, Chen, Yuan, and Guo]{gu2022vector}
Shuyang Gu, Dong Chen, Jianmin Bao, Fang Wen, Bo~Zhang, Dongdong Chen, Lu~Yuan, and Baining Guo.
\newblock Vector quantized diffusion model for text-to-image synthesis.
\newblock In \emph{Proceedings of the IEEE/CVF conference on computer vision and pattern recognition}, pp.\  10696--10706, 2022.

\bibitem[Heusel et~al.(2017)Heusel, Ramsauer, Unterthiner, Nessler, and Hochreiter]{DBLP:conf/nips/HeuselRUNH17}
Martin Heusel, Hubert Ramsauer, Thomas Unterthiner, Bernhard Nessler, and Sepp Hochreiter.
\newblock Gans trained by a two time-scale update rule converge to a local nash equilibrium.
\newblock In \emph{NIPS}, 2017.

\bibitem[Ho \& Salimans(2022)Ho and Salimans]{ho2022classifier}
Jonathan Ho and Tim Salimans.
\newblock Classifier-free diffusion guidance.
\newblock \emph{arXiv preprint arXiv:2207.12598}, 2022.

\bibitem[Ho et~al.(2020)Ho, Jain, and Abbeel]{ho2020denoising}
Jonathan Ho, Ajay Jain, and Pieter Abbeel.
\newblock Denoising diffusion probabilistic models.
\newblock \emph{Advances in Neural Information Processing Systems}, 33, 2020.

\bibitem[Hou et~al.(2022)Hou, Zhang, Li, and Shen]{hou2022textface}
Xianxu Hou, Xiaokang Zhang, Yudong Li, and Linlin Shen.
\newblock Textface: Text-to-style mapping based face generation and manipulation.
\newblock \emph{IEEE Transactions on Multimedia}, 2022.

\bibitem[Hyv{\"a}rinen(2005)]{hyvarinen2005estimation}
Aapo Hyv{\"a}rinen.
\newblock Estimation of non-normalized statistical models by score matching.
\newblock \emph{Journal of Machine Learning Research}, 6\penalty0 (Apr):\penalty0 695--709, 2005.

\bibitem[Krizhevsky et~al.(2012)Krizhevsky, Sutskever, and Hinton]{krizhevsky2012imagenet}
Alex Krizhevsky, Ilya Sutskever, and Geoffrey~E Hinton.
\newblock Imagenet classification with deep convolutional neural networks.
\newblock \emph{Advances in neural information processing systems}, 25, 2012.

\bibitem[LeCun et~al.(2015)LeCun, Bengio, and Hinton]{lecun2015deep}
Yann LeCun, Yoshua Bengio, and Geoffrey Hinton.
\newblock Deep learning.
\newblock \emph{nature}, 521\penalty0 (7553):\penalty0 436--444, 2015.

\bibitem[Li et~al.(2022)Li, Torr, and Lukasiewicz]{li2022memory}
Bowen Li, Philip~HS Torr, and Thomas Lukasiewicz.
\newblock Memory-driven text-to-image generation.
\newblock \emph{arXiv preprint arXiv:2208.07022}, 2022.

\bibitem[Liu et~al.(2023)Liu, Vermeulen, Fitzmaurice, and Matejka]{liu20233dall}
Vivian Liu, Jo~Vermeulen, George Fitzmaurice, and Justin Matejka.
\newblock 3dall-e: Integrating text-to-image ai in 3d design workflows.
\newblock In \emph{Proceedings of the 2023 ACM designing interactive systems conference}, pp.\  1955--1977, 2023.

\bibitem[Lloyd(1982)]{lloyd1982least}
Stuart Lloyd.
\newblock Least squares quantization in pcm.
\newblock \emph{IEEE transactions on information theory}, 28\penalty0 (2):\penalty0 129--137, 1982.

\bibitem[Lukasik et~al.(2020)Lukasik, Bhojanapalli, Menon, and Kumar]{lukasik2020does}
Michal Lukasik, Srinadh Bhojanapalli, Aditya Menon, and Sanjiv Kumar.
\newblock Does label smoothing mitigate label noise?
\newblock In \emph{International Conference on Machine Learning}, pp.\  6448--6458. PMLR, 2020.

\bibitem[Mirza \& Osindero(2014)Mirza and Osindero]{DBLP:journals/corr/MirzaO14}
Mehdi Mirza and Simon Osindero.
\newblock Conditional generative adversarial nets.
\newblock \emph{CoRR}, abs/1411.1784, 2014.
\newblock URL \url{http://arxiv.org/abs/1411.1784}.

\bibitem[M{\"u}ller et~al.(2019)M{\"u}ller, Kornblith, and Hinton]{muller2019does}
Rafael M{\"u}ller, Simon Kornblith, and Geoffrey~E Hinton.
\newblock When does label smoothing help?
\newblock \emph{Advances in neural information processing systems}, 32, 2019.

\bibitem[Peebles \& Xie(2023)Peebles and Xie]{peebles2023scalable}
William Peebles and Saining Xie.
\newblock Scalable diffusion models with transformers.
\newblock In \emph{Proceedings of the IEEE/CVF International Conference on Computer Vision}, pp.\  4195--4205, 2023.

\bibitem[Peng et~al.(2021)Peng, Zhou, Sun, Cao, Wu, Huang, and Ji]{peng2021knowledge}
Jun Peng, Yiyi Zhou, Xiaoshuai Sun, Liujuan Cao, Yongjian Wu, Feiyue Huang, and Rongrong Ji.
\newblock Knowledge-driven generative adversarial network for text-to-image synthesis.
\newblock \emph{IEEE Transactions on Multimedia}, 2021.

\bibitem[Radford et~al.(2021)Radford, Kim, Hallacy, Ramesh, Goh, Agarwal, Sastry, Askell, Mishkin, Clark, et~al.]{radford2021learning}
Alec Radford, Jong~Wook Kim, Chris Hallacy, Aditya Ramesh, Gabriel Goh, Sandhini Agarwal, Girish Sastry, Amanda Askell, Pamela Mishkin, Jack Clark, et~al.
\newblock Learning transferable visual models from natural language supervision.
\newblock In \emph{International conference on machine learning}, pp.\  8748--8763. PMLR, 2021.

\bibitem[Ramesh et~al.(2022)Ramesh, Dhariwal, Nichol, Chu, and Chen]{ramesh2022hierarchical}
Aditya Ramesh, Prafulla Dhariwal, Alex Nichol, Casey Chu, and Mark Chen.
\newblock Hierarchical text-conditional image generation with clip latents.
\newblock \emph{arXiv preprint arXiv:2204.06125}, 1\penalty0 (2):\penalty0 3, 2022.

\bibitem[Reed et~al.(2016{\natexlab{a}})Reed, Akata, Mohan, Tenka, Schiele, and Lee]{reed2016learning}
Scott~E Reed, Zeynep Akata, Santosh Mohan, Samuel Tenka, Bernt Schiele, and Honglak Lee.
\newblock Learning what and where to draw.
\newblock In \emph{NIPS}, 2016{\natexlab{a}}.

\bibitem[Reed et~al.(2016{\natexlab{b}})Reed, Akata, Yan, Logeswaran, Schiele, and Lee]{reed2016generative}
Scott~E Reed, Zeynep Akata, Xinchen Yan, Lajanugen Logeswaran, Bernt Schiele, and Honglak Lee.
\newblock Generative adversarial text to image synthesis.
\newblock \emph{international conference on machine learning}, pp.\  1060--1069, 2016{\natexlab{b}}.

\bibitem[Rombach et~al.(2022)Rombach, Blattmann, Lorenz, Esser, and Ommer]{rombach2022high}
Robin Rombach, Andreas Blattmann, Dominik Lorenz, Patrick Esser, and Bj{\"o}rn Ommer.
\newblock High-resolution image synthesis with latent diffusion models.
\newblock In \emph{Proceedings of the IEEE/CVF conference on computer vision and pattern recognition}, pp.\  10684--10695, 2022.

\bibitem[Ruan et~al.(2021)Ruan, Zhang, Zhang, Fan, Tang, Liu, and Chen]{DBLP:journals/corr/abs-2108-12141}
Shulan Ruan, Yong Zhang, Kun Zhang, Yanbo Fan, Fan Tang, Qi~Liu, and Enhong Chen.
\newblock {DAE-GAN:} dynamic aspect-aware {GAN} for text-to-image synthesis.
\newblock In \emph{2021 {IEEE/CVF} International Conference on Computer Vision, {ICCV} 2021, Montreal, QC, Canada, October 10-17, 2021}, pp.\  13940--13949. {IEEE}, 2021.
\newblock \doi{10.1109/ICCV48922.2021.01370}.
\newblock URL \url{https://doi.org/10.1109/ICCV48922.2021.01370}.

\bibitem[Saharia et~al.(2022)Saharia, Chan, Saxena, Li, Whang, Denton, Ghasemipour, Ayan, Mahdavi, Lopes, et~al.]{saharia2022photorealistic}
Chitwan Saharia, William Chan, Saurabh Saxena, Lala Li, Jay Whang, Emily Denton, Seyed Kamyar~Seyed Ghasemipour, Burcu~Karagol Ayan, S~Sara Mahdavi, Rapha~Gontijo Lopes, et~al.
\newblock Photorealistic text-to-image diffusion models with deep language understanding.
\newblock \emph{arXiv preprint arXiv:2205.11487}, 2022.

\bibitem[Salimans et~al.(2016)Salimans, Goodfellow, Zaremba, Cheung, Radford, and Chen]{DBLP:conf/nips/SalimansGZCRCC16}
Tim Salimans, Ian~J. Goodfellow, Wojciech Zaremba, Vicki Cheung, Alec Radford, and Xi~Chen.
\newblock Improved techniques for training gans.
\newblock In \emph{NIPS}, pp.\  2226--2234, 2016.

\bibitem[Sauer et~al.(2023)Sauer, Karras, Laine, Geiger, and Aila]{sauer2023stylegan}
Axel Sauer, Tero Karras, Samuli Laine, Andreas Geiger, and Timo Aila.
\newblock Stylegan-t: Unlocking the power of gans for fast large-scale text-to-image synthesis.
\newblock In \emph{International conference on machine learning}, pp.\  30105--30118. PMLR, 2023.

\bibitem[Sheynin et~al.(2022)Sheynin, Ashual, Polyak, Singer, Gafni, Nachmani, and Taigman]{sheynin2022knn}
Shelly Sheynin, Oron Ashual, Adam Polyak, Uriel Singer, Oran Gafni, Eliya Nachmani, and Yaniv Taigman.
\newblock Knn-diffusion: Image generation via large-scale retrieval.
\newblock \emph{arXiv preprint arXiv:2204.02849}, 2022.

\bibitem[Song \& Ermon(2019)Song and Ermon]{DBLP:conf/nips/SongE19}
Yang Song and Stefano Ermon.
\newblock Generative modeling by estimating gradients of the data distribution.
\newblock In Hanna~M. Wallach, Hugo Larochelle, Alina Beygelzimer, Florence d'Alch{\'{e}}{-}Buc, Emily~B. Fox, and Roman Garnett (eds.), \emph{Advances in Neural Information Processing Systems}, pp.\  11895--11907, 2019.

\bibitem[Song et~al.(2021)Song, Sohl{-}Dickstein, Kingma, Kumar, Ermon, and Poole]{DBLP:conf/iclr/0011SKKEP21}
Yang Song, Jascha Sohl{-}Dickstein, Diederik~P. Kingma, Abhishek Kumar, Stefano Ermon, and Ben Poole.
\newblock Score-based generative modeling through stochastic differential equations.
\newblock In \emph{9th International Conference on Learning Representations, {ICLR}}. OpenReview.net, 2021.

\bibitem[Tan et~al.(2022)Tan, Liu, Yin, and Li]{DBLP:journals/tmm/TanLYL22}
Hongchen Tan, Xiuping Liu, Baocai Yin, and Xin Li.
\newblock Cross-modal semantic matching generative adversarial networks for text-to-image synthesis.
\newblock \emph{{IEEE} Trans. Multim.}, 2022.

\bibitem[Tao et~al.(2022)Tao, Tang, Wu, Jing, Bao, and Xu]{DBLP:journals/corr/abs-2008-05865}
Ming Tao, Hao Tang, Fei Wu, Xiaoyuan Jing, Bing{-}Kun Bao, and Changsheng Xu.
\newblock {DF-GAN:} {A} simple and effective baseline for text-to-image synthesis.
\newblock In \emph{{IEEE/CVF} Conference on Computer Vision and Pattern Recognition, {CVPR} 2022, New Orleans, LA, USA, June 18-24, 2022}, pp.\  16494--16504. {IEEE}, 2022.
\newblock \doi{10.1109/CVPR52688.2022.01602}.
\newblock URL \url{https://doi.org/10.1109/CVPR52688.2022.01602}.

\bibitem[Tao et~al.(2023)Tao, Bao, Tang, and Xu]{tao2023galip}
Ming Tao, Bing-Kun Bao, Hao Tang, and Changsheng Xu.
\newblock Galip: Generative adversarial clips for text-to-image synthesis.
\newblock In \emph{Proceedings of the IEEE/CVF Conference on Computer Vision and Pattern Recognition}, pp.\  14214--14223, 2023.

\bibitem[Vedaldi \& Zisserman(2016)Vedaldi and Zisserman]{vedaldi2016vgg}
Andrea Vedaldi and Andrew Zisserman.
\newblock Vgg convolutional neural networks practical.
\newblock \emph{Department of Engineering Science, University of Oxford}, 66, 2016.

\bibitem[Xu et~al.(2018)Xu, Zhang, Huang, Zhang, Gan, Huang, and He]{DBLP:conf/cvpr/XuZHZGH018}
Tao Xu, Pengchuan Zhang, Qiuyuan Huang, Han Zhang, Zhe Gan, Xiaolei Huang, and Xiaodong He.
\newblock Attngan: Fine-grained text to image generation with attentional generative adversarial networks.
\newblock In \emph{CVPR}. Computer Vision Foundation / {IEEE} Computer Society, 2018.

\bibitem[Ye \& Liu(2024)Ye and Liu]{ye2024score}
Senmao Ye and Fei Liu.
\newblock Score mismatching for generative modeling.
\newblock \emph{Neural Networks}, 175:\penalty0 106311, 2024.

\bibitem[Ye et~al.(2023)Ye, Wang, Tan, and Liu]{ye2023recurrent}
Senmao Ye, Huan Wang, Mingkui Tan, and Fei Liu.
\newblock Recurrent affine transformation for text-to-image synthesis.
\newblock \emph{IEEE Transactions on Multimedia}, 2023.

\bibitem[Yin \& Li(2023)Yin and Li]{yin2023systematic}
Ming-Yue Yin and Jian-Guang Li.
\newblock A systematic review on digital human models in assembly process planning.
\newblock \emph{The International Journal of Advanced Manufacturing Technology}, 125\penalty0 (3):\penalty0 1037--1059, 2023.

\bibitem[Yu et~al.(2022)Yu, Xu, Koh, Luong, Baid, Wang, Vasudevan, Ku, Yang, Ayan, et~al.]{yu2022scaling}
Jiahui Yu, Yuanzhong Xu, Jing~Yu Koh, Thang Luong, Gunjan Baid, Zirui Wang, Vijay Vasudevan, Alexander Ku, Yinfei Yang, Burcu~Karagol Ayan, et~al.
\newblock Scaling autoregressive models for content-rich text-to-image generation.
\newblock \emph{arXiv preprint arXiv:2206.10789}, 2\penalty0 (3):\penalty0 5, 2022.

\bibitem[Zhang et~al.(2019)Zhang, Xu, Li, Zhang, Wang, Huang, and Metaxas]{DBLP:journals/pami/ZhangXLZWHM19}
Han Zhang, Tao Xu, Hongsheng Li, Shaoting Zhang, Xiaogang Wang, Xiaolei Huang, and Dimitris~N. Metaxas.
\newblock Stackgan++: Realistic image synthesis with stacked generative adversarial networks.
\newblock \emph{{IEEE} Trans. Pattern Anal. Mach. Intell.}, 41\penalty0 (8):\penalty0 1947--1962, 2019.

\bibitem[Zhang et~al.(2017)Zhang, Cisse, Dauphin, and Lopez-Paz]{zhang2017mixup}
Hongyi Zhang, Moustapha Cisse, Yann~N Dauphin, and David Lopez-Paz.
\newblock mixup: Beyond empirical risk minimization.
\newblock \emph{arXiv preprint arXiv:1710.09412}, 2017.

\bibitem[Zhang \& Schomaker(2021)Zhang and Schomaker]{DBLP:conf/ijcnn/ZhangS21}
Zhenxing Zhang and Lambert Schomaker.
\newblock {DTGAN:} dual attention generative adversarial networks for text-to-image generation.
\newblock In \emph{IJCNN}, pp.\  1--8. {IEEE}, 2021.

\bibitem[Zhong et~al.(2020)Zhong, Zheng, Kang, Li, and Yang]{zhong2020random}
Zhun Zhong, Liang Zheng, Guoliang Kang, Shaozi Li, and Yi~Yang.
\newblock Random erasing data augmentation.
\newblock In \emph{Proceedings of the AAAI conference on artificial intelligence}, volume~34, pp.\  13001--13008, 2020.

\bibitem[Zhu et~al.(2019)Zhu, Pan, Chen, and Yang]{DBLP:conf/cvpr/ZhuP0019}
Minfeng Zhu, Pingbo Pan, Wei Chen, and Yi~Yang.
\newblock {DM-GAN:} dynamic memory generative adversarial networks for text-to-image synthesis.
\newblock In \emph{CVPR}, 2019.

\end{thebibliography}
\bibliographystyle{iclr2025_conference}

\appendix
\begin{figure*}[h]
	\centering
	\includegraphics[width = \textwidth]{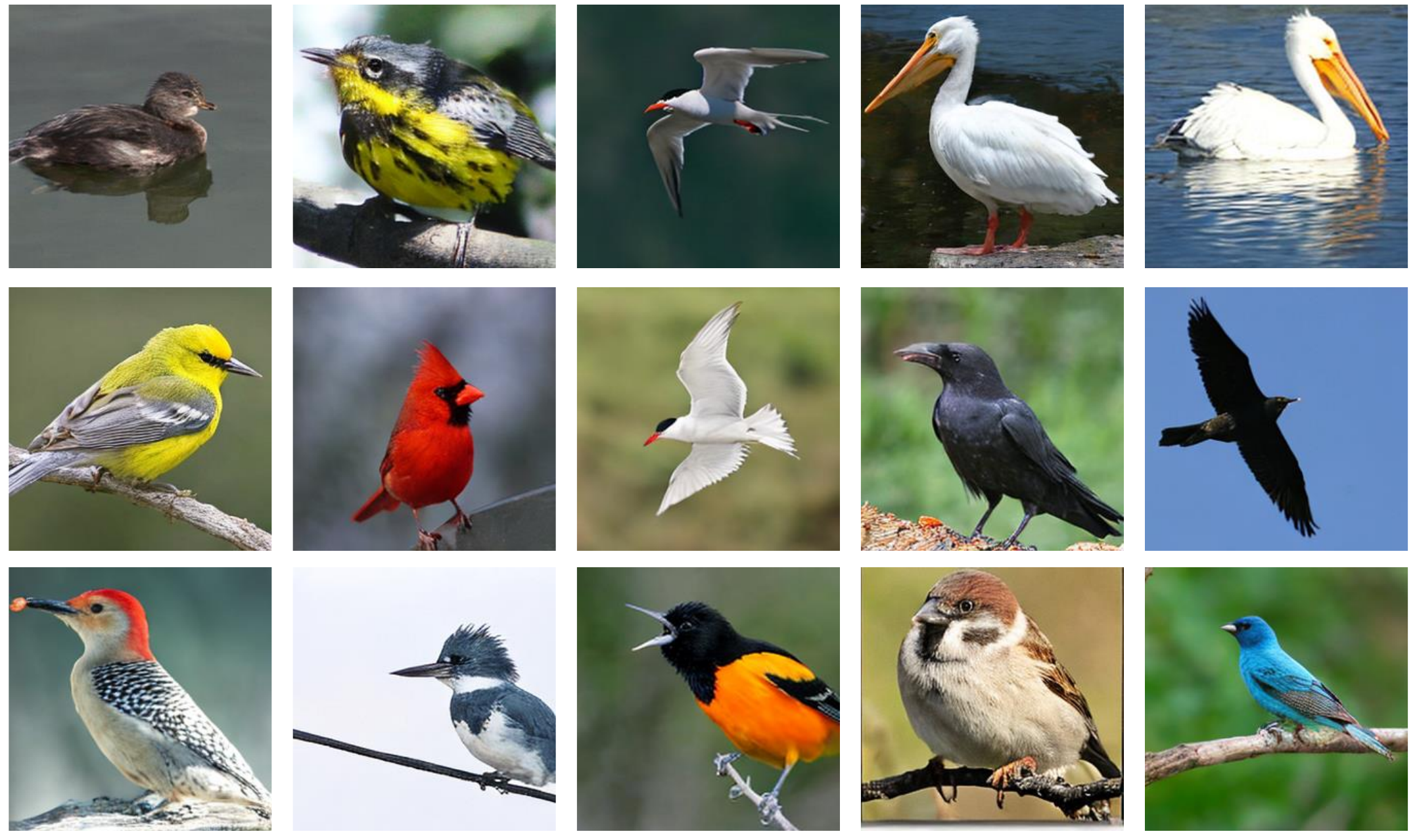}
	\caption{\label{Fig:rat} More generated images from the CUB dataset. The text descriptions are omitted for simplicity.}
	\label{Fig:framework}
\end{figure*}
\begin{figure*}[t]
	\centering
	\includegraphics[width = \textwidth]{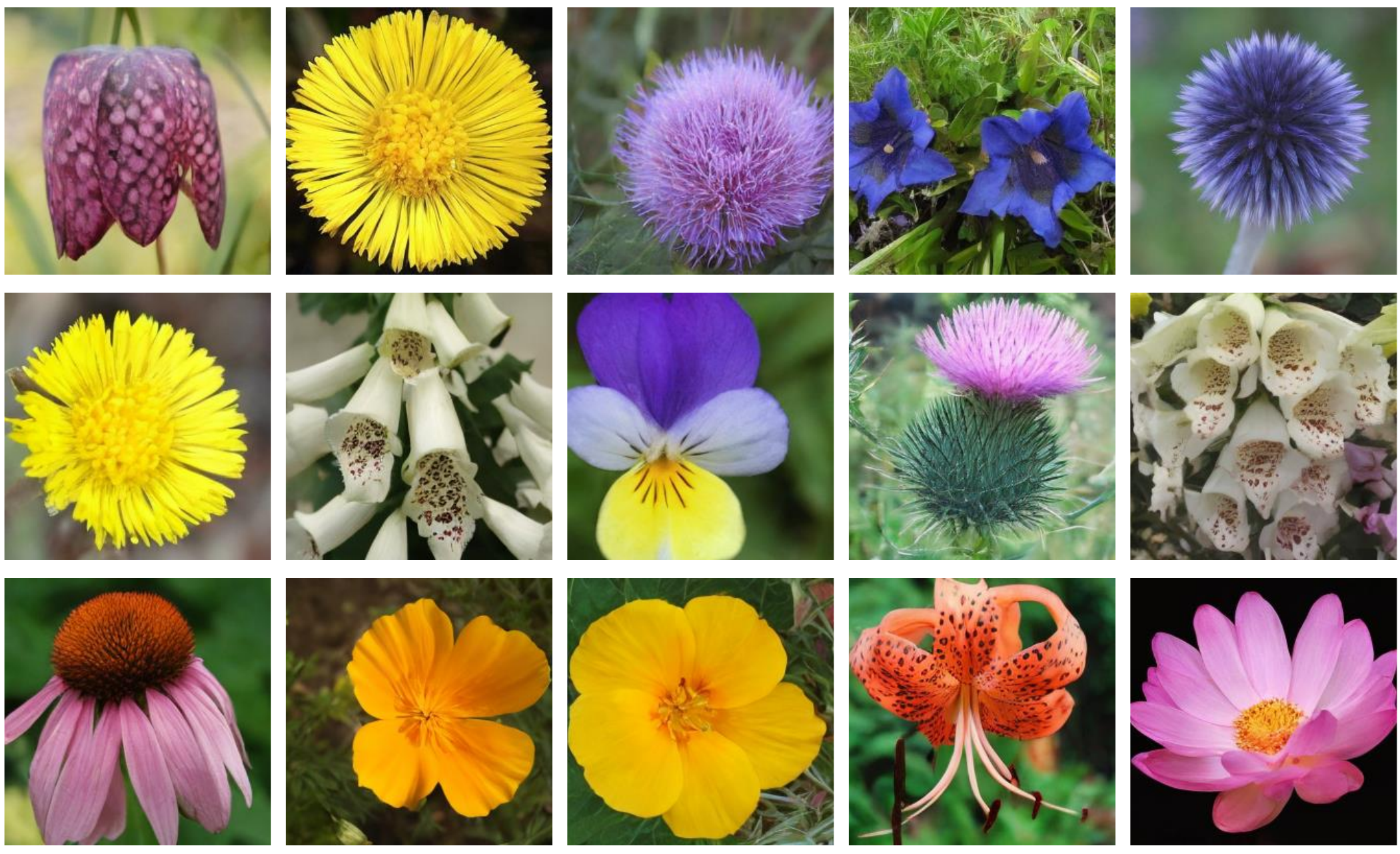}
	\caption{\label{Fig:rat} More generated images from the Oxford dataset. The text descriptions are omitted for simplicity.}
	\label{Fig:framework}
\end{figure*}
\begin{figure*}[t]
	\centering
	\includegraphics[width = \textwidth]{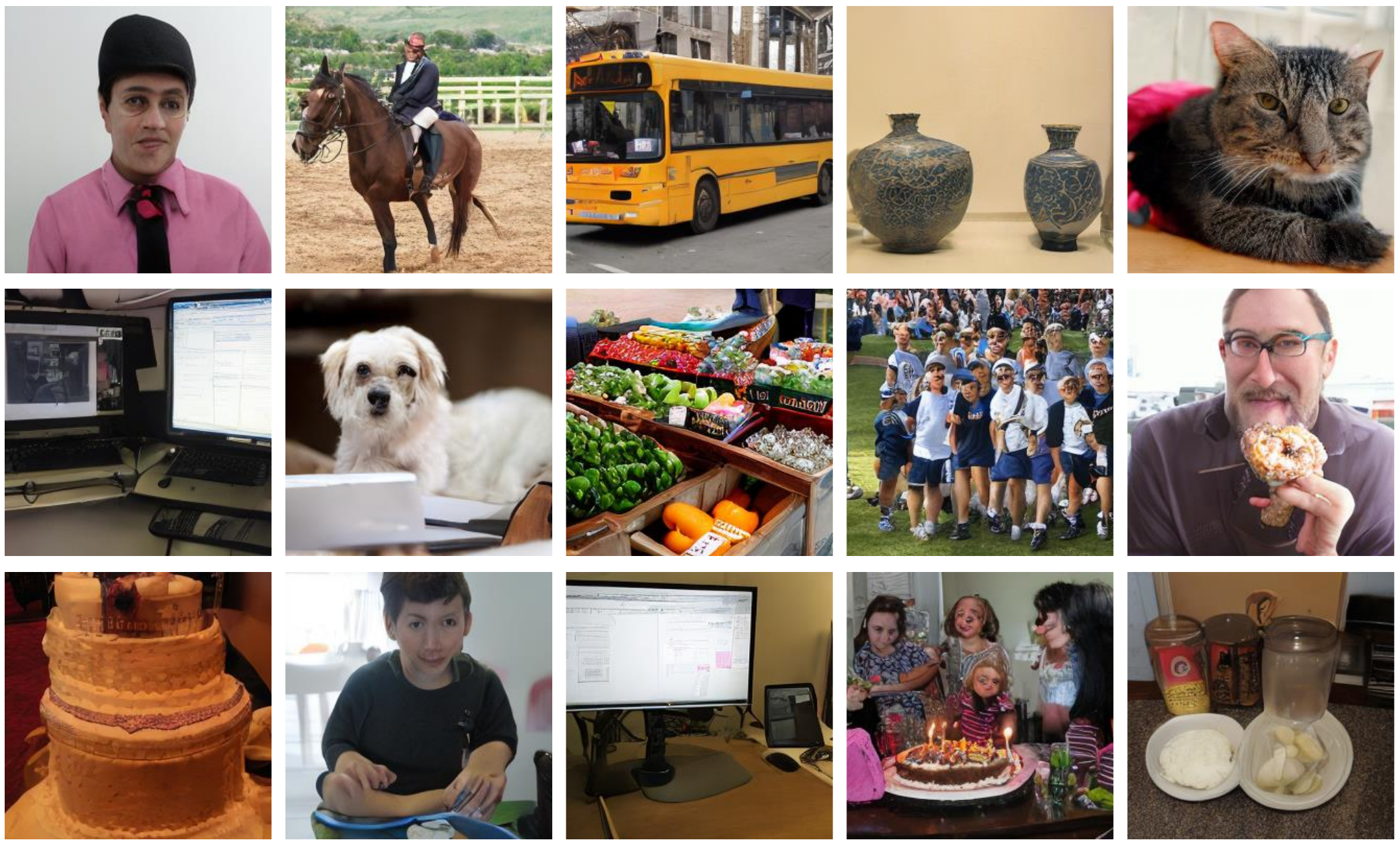}
	\caption{\label{Fig:rat} More generated images from the COCO dataset. The text descriptions are omitted for simplicity.}
	\label{Fig:framework}
\end{figure*}
\end{document}


\maketitle

\appendix
\begin{figure*}[h]
	\centering
	\includegraphics[width = \textwidth]{ICLR 2025 Template/fig/bird.pdf}
	\caption{\label{Fig:rat} More generated images from the CUB dataset. The text descriptions are omitted for simplicity.}
	\label{Fig:framework}
\end{figure*}
\begin{figure*}[t]
	\centering
	\includegraphics[width = \textwidth]{ICLR 2025 Template/fig/flower.pdf}
	\caption{\label{Fig:rat} More generated images from the Oxford dataset. The text descriptions are omitted for simplicity.}
	\label{Fig:framework}
\end{figure*}
\begin{figure*}[t]
	\centering
	\includegraphics[width = \textwidth]{ICLR 2025 Template/fig/coco.pdf}
	\caption{\label{Fig:rat} More generated images from the COCO dataset. The text descriptions are omitted for simplicity.}
	\label{Fig:framework}
\end{figure*}